\documentclass{article}
\usepackage{arabtex}
\usepackage{utf8}
\setcode{utf8}
\usepackage{arxiv}
\usepackage[utf8]{inputenc} 
\usepackage[T1]{fontenc}    
\usepackage{hyperref}       
\usepackage{url}            
\usepackage{booktabs}       
\usepackage{amsfonts}       
\usepackage{nicefrac}       
\usepackage{microtype}      
\usepackage{lipsum}		
\usepackage{graphicx}
\usepackage{natbib}
\usepackage{doi}

\title{Hate speech detection in Algerian dialect using deep learning}

\date{} 			

\author{
    \href{https://orcid.org/0000-0002-3794-844X}{\includegraphics[scale=0.06]{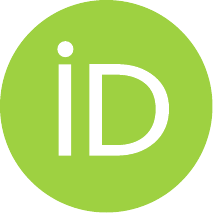}\hspace{1mm}Dihia LANASRI}\thanks{Dihia Lanasri, Juan Olano, and Sifal Klioui contributed equally to this work.} \\
    OMDENA\\
    New York, USA\\	
    \texttt{dihia.lanasri@gmail.com} \\
    \And
    Juan OLANO\footnotemark[1] \\
    OMDENA\\
    New York, USA\\	
    \texttt{juan\_olano@yahoo.com}  \\
    \And
    \href{https://orcid.org/0009-0000-4432-7572}{\includegraphics[scale=0.06]{orcid.pdf}\hspace{1mm}Sifal KLIOUI}\footnotemark[1] \\
    OMDENA\\
    New York, USA\\	
    \texttt{sifal.klioui@gmail.com}  \\
    \And
    Sin Liang Lee \\
    OMDENA\\
    New York, USA\\	
    \texttt{mangojamlee@gmail.com}  \\
    \And
    \href{https://orcid.org/0000-0003-1414-6425}{\includegraphics[scale=0.06]{orcid.pdf}\hspace{1mm}Lamia SEKKAI} \\
    OMDENA\\
    New York, USA\\	
    \texttt{lsekkai@gmail.com}  \\ 
}

\renewcommand{\shorttitle}{Hate Speech Detection in Algerian Dialect}

\hypersetup{
pdftitle={Hate Speech Detection in Algerian Dialect},
pdfsubject={NLP, AI},
pdfauthor={D.LANASRI, J.OLANO, S.KLIOUI, S.L.LEE, L.SEKKAI},
pdfkeywords={Hate Speech, Algerian dialect, Deep Learning, DziriBERT, FastText},
}

\begin{document}
\maketitle

\begin{abstract}
With the proliferation of hate speech on social networks under different formats, such as abusive language, cyberbullying, and violence, etc., people have experienced a significant increase in violence, putting them in uncomfortable situations and threats. Plenty of efforts have been dedicated in the last few years to overcome this phenomenon to detect hate speech in different structured languages like English, French, Arabic, and others. However, a reduced number of works deal with Arabic dialects like Tunisian, Egyptian, and Gulf, mainly the Algerian ones. To fill in the gap, we propose in this work a complete approach for detecting hate speech on online Algerian messages. Many deep learning architectures have been evaluated on the corpus we created from some Algerian social networks (Facebook, YouTube, and Twitter). This corpus contains more than 13.5K documents in Algerian dialect written in Arabic, labeled as hateful or non-hateful. Promising results are obtained, which show the efficiency of our approach. 
\end{abstract}

\keywords{Hate Speech \and Algerian dialect \and Deep Learning \and DziriBERT \and FastText}

\section{Introduction}
Hate speech detection, or detection of offensive messages in social networks, communication forums, and websites, is an exciting and hot research topic. Many hate crimes and attacks in our current life started from social network posts and comments \cite{macavaney2019hate}. Studying this phenomenon is imperative for online communities to keep a safe environment for their users. It also has a significant benefit for security authorities and states to ensure the safety of citizens and prevent crimes and attacks.

A universally accepted definition of hate speech is currently unavailable \cite{bogdani2021beyond} because of the variation of cultures, societies, and local languages. Other difficulties include the diversity of national laws, the variety of online communities, and forms of online hate speech. Various definitions are proposed.

According to the Encyclopedia of the American Constitution: "Hate speech is speech that attacks a person or group based on attributes such as race, religion, ethnic origin, national origin, sex, disability, sexual orientation, or gender identity." \cite{nockleby2000hate}. Today, many authors largely used this definition \cite{guellil2022ara}. Facebook considers hate speech as "a direct attack on people based on protected characteristics—race, ethnicity, national origin, religious affiliation, sexual orientation, caste, sex, gender, gender identity, and serious disease or disability. We also provide some protections for immigration status." \footnote{Community Standards; Available on:\url{https://www.facebook.com/communitystandards/objectionable_content}}. Davidson et al., who defines hate speech as "language that is used to express hatred towards a targeted group or is intended to be derogatory, to humiliate, or to insult the members of the group" propose one of the most accepted definitions \cite{davidson2017automated}. Alternatively, the one proposed by Fortuna et al., "Hate speech is a language that attacks or diminishes, that incites violence or hate against groups, based on specific characteristics such as physical appearance, religion, descent, national or ethnic origin, sexual orientation, gender identity or other, and it can occur with different linguistic styles, even in subtle forms or when humor is used." \cite{fortuna2018survey}.

The literature review shows that the term \textit{Hate speech} (which is the most commonly used) has various synonym terms such as abusive speech, offensive language, cyberbullying, or sexism detection \cite{schmidt2017survey}. Many works have been published in the context of hate speech detection for different standard and structured languages, like French \cite{battistelli2020building}, English \cite{alkomah2022literature}, Spanish \cite{plaza2021comparing}, and Arabic \cite{albadi2018they}. These languages are known for their standardization with well-known grammar and structure, which make the language processing well mastered. However, detecting hate speech in dialects, mainly Arabic ones such as Libyan, Egyptian, and Iraqi, etc. is still challenging and complex work \cite{mulki2019hsab}. Even if they are derived from the literal Arabic language, each country's specific vocabulary and semantics are added or defined.

In this work, we are interested in detecting hate speech in the Algerian dialect. This latter is one of the complex dialects \cite{mezzoudj2019arabic} characterized by the variety of its sub-dialects according to each region within the country. Algeria is a country with 58 regions; each one has a specificity in its spoken language with different words and meanings. The same word may have various meanings for each region; for example, '\textit{Flouka}' in the east means '\textit{earrings}.' In the north, it means '\textit{small boat}'.

Moreover, new '\textit{odd}' words are continually added to the Algerian vocabulary. The Algerian dialect is known for its morphological and orthographic richness. Facing this situation, treating and understanding the Algerian dialect for hate speech detection is a complex work. The importance of this project for the Algerian context encourages us to work on this problem.

To the best of our knowledge, only few works have been proposed for hate speech detection in the Algerian dialect \cite{boucherit2022offensive,menifi2022transfer}. Some other related topics are treated like sentiment analysis \cite{abdelli2019sentiment}, sexism detection \cite{guellil2021sexism} which may be exploited to analyze the hate speech.

In this paper, we proposed a complete end-to-end natural language processing (NLP) approach for hate speech detection in the Algerian dialect. Our approach covers the main steps of an NLP project, including data collection, data annotation, feature extraction, and then model development based on machine and deep learning, model evaluation, and inference.

Moreover, we have evaluated various machine and deep learning architectures on our corpus built from diverse social networks (YouTube, Twitter, and Facebook) for several years (between 2017 and 2023). This corpus contains more than 13.5K annotated documents in Algerian dialect written in Arabic characters. Two classes are used for annotation (hateful, non-hateful). This work allows us essentially to provide a wealthy evaluation of many deep learning architectures, an essential value for academic and industrial communities. The obtained results are promising, and continuous tests are performed for further results.

This paper is structured as follows: Section 2 presents a necessary background, Section 3 reviews the most important related works, Section 4 details our proposed approach and evaluated models, Section 5 discusses the obtained results, and Section 6 concludes the paper.

\section{Background}
\label{sec:background}
Hate speech is commonly defined as any communication that disparages a target group of people based on some characteristic such as race, color, ethnicity, gender, sexual orientation, nationality, religion, or other characteristic \cite{de2018hate}. 

\subsection{Hate speech}
According to \cite{al2019detection} hate speech is categorized into five categories: (1) gendered hate speech, including any form of misogyny and sexism; (2) religious hate speech including any religious discrimination, such as Islamic sects, anti-Christian, etc.; (3) racist hate speech including any racial offense or tribalism, and xenophobia; (4) disability including any sort of offense to an individual suffering from health problems; and (5) political hate speech can refer to any abuse and offense against politicians \cite{guellil2022ara}.

\subsection{Algerian Dialect and Arabic Languages} Arabic is the official language of 25 countries\footnote{\url{https://worldpopulationreview.com/country-rankings/arabic-speaking-countries}}. More than 400 million people around the world speak this language. Arabic is also recognized as the 4th most-used language on the Internet \cite{boudad2018sentiment}. Arabic is classified into three categories \cite{habash2022introduction}: (1) Classical Arabic (CA), which is the form of the Arabic language used in literary texts. The Quran is considered the highest form of CA text \cite{sharaf2012qurana}. (2) Modern Standard Arabic (MSA) is used for writing and formal conversations. (3) Dialectal Arabic is used in daily life communication, informal exchanges, etc. \cite{boudad2018sentiment} like the Algerian dialect, Tunisian dialect, etc.
\\
The Algerian dialect on social networks can be written with Arabic characters (\RL{كرهت من هادي الميزيرية}), Latin characters (kreht men hadi el miziria), or a mix of them. This dialect does not respect a specific syntax or grammar; the same word may have many meanings according to each region in Algeria (having 58 regions). In addition, the same word may be written in different manners (exp. Inchallah , nchallah, n'shala, etc. to say hopefully). This paper does not deal with the Amazigh (Berber) language, another language spoken and written in Algeria, which is entirely different from the Algerian dialect. It has its own vocabulary and grammar.

\section{Related Work}
A restricted number of works have been published in the context of hate speech detection dealing with Arabic dialects. This section will analyze the most important NLP approaches dedicated to (1) the Algerian dialect and (2) Other Arabic dialects like Iraqi, Egyptian, Syrian, and Tunisian. This analysis helps us to identify the used approaches, models, and corpora.

\subsection{Hate speech detection in Algerian dialect}
\cite{guellil2022ara,guellil2021sexism} Developed the first approach and corpus in Algerian dialect for hate speech detection against women in Arabic community on social media. This corpus contains more than 373K YouTube comments. Two different algorithms for feature extraction were used: Word2vec with machine learning models (GaussianNB, LogisticRegression, RandomForset, SGDClassifier, and LinearSVC) and FastText with Deep learning models (deep Convolutional Neural Network (CNN), long short-term memory (LSTM) network and Bi-directional LSTM (BiLSTM) network). Simulation results demonstrated the best performance of the CNN model with FastText.

\cite{boucherit2022offensive} addressed the problem of detecting offensive and abusive content in Facebook comments. The corpus contains 8.7K comments in Algerian dialect written in Arabic and Latin characters, manually annotated as usual, abusive, and offensive. They used BiLSTM, CNN, FastText, SVM, and Multinomial Naive Bayes (NB) as classifiers. The experimental results showed that SVM and Multinomial NB classifiers outperformed all the other classifiers.

\cite{abainia2022new} addressed the offensive language detection in the Amazigh language, which is one of the under-resourced languages. They were interested in the Kabyle dialect. A new corpus of offensive Amazigh language is proposed containing 6.2K documents collected from Facebook and manually annotated as usual or offensive. A new lexicon of offensive and abusive Amazigh words with 12.6k entries is also developed. Many models have been evaluated, like SVM and Multinomial Naive Bayes classifiers tested with tf-idf. FastText was tested with deep learning models CNN and BiLSTM. The naive statistical classifier based on lexicon checking was the winner classifier.

\cite{mazari2023deep} introduced a new dataset for Algerian dialect toxic text detection. An annotated multi-label dataset is built, containing around 14K comments extracted from Facebook, YouTube, and Twitter and labeled as hate speech, offensive language, and cyberbullying. Several tests have been conducted using many classification models of traditional machine learning: Random Forest, Naıve Bayes, Linear Support Vector (LSV), Stochastic Gradient Descent (SGD), and Logistic Regression. Furthermore, several assessments have been conducted using Deep Learning models such as CNN, LSTM, Gated Recurrent Unit (GRU), BiLSTM and Bidirectional-GRU (Bi-GRU). Results demonstrated the best performance of LSV, BiLSTM, and MLP when associated with the SGD model.

\cite{guellil2020detecting} proposed a system for detecting hateful speech in Arabic political debates. The approach was evaluated against a hateful corpus concerning Algerian political debates. It contains 5K YouTube comments in MSA and Algerian dialects, written in both Arabic and Latin characters. Both classical algorithms of classification (Gaussian NB, Logistic Regression, Random Forest, SGD Classifier, and Linear SVC(LSVC)) and deep learning algorithms (CNN, multilayer perceptron (MLP), LSTM, and BiLSTM) are tested. For extracting features, the authors use Word2vec and FastText with their two implementations, namely, Skip Gram and CBOW. Simulation results demonstrate the best performance of LSVC, BiLSTM and MLP.

\cite{mohdeb2022evaluating} proposed an approach for analysis and the detection of dialectal Arabic hate speech that targeted African refugees and illegal migrants on the YouTube Algerian space. The corpus contains more than 4K comments annotated as Incitement, Hate, Refusing with non-hateful words, Sympathetic, and Comment. The transfer learning approach has been exploited for classification. The experiments show that the AraBERT monolingual transformer outperforms the mono-dialectal transformer DziriBERT and the cross-lingual transformers mBERT and XLM-R.

\subsection{Hate speech detection in other Arabic dialects} 
Various datasets or corpora were published in different dialects, which can be used for different purposes like hate speech, racism, violence, etc. detection.

\cite{albayari2022instagram} is the first work to propose a corpus built from Instagram comments. This corpus contains 198K comments, written in MSA and three different dialects: Egyptian, Gulf, and Levantine. The comments were annotated as neutral, toxic, and Bullying. 
\cite{al2018optimized} and \cite{haidar2019arabic} datasets are collected from Twitter containing respectively 20K and 34K multi-dialectal Arabic tweets annotated as bullying and non-bullying labels. These tweets were from various dialects (Lebanon, Egypt, and the Gulf area). Moreover, two other datasets were proposed by \cite{mubarak2017abusive}. The first one with 1.1K tweets in different dialects and the second dataset contains 32K inappropriate comments collected from a famous Arabic news site and annotated as obscene, offensive, or clean. 
\cite{albadi2018they} proposed the religious hate speech detection where a multi-dialectal dataset of 6.6K tweets was introduced. It included an identification of the religious groups targeted by hate speech.
\cite{alakrot2018dataset} also provided a dataset of 16K Egyptian, Iraqi, and Libyan comments collected from YouTube. The comments were annotated as either offensive, inoffensive, or neutral.

T-HSAB \cite{haddad2019t} and L-HSAB \cite{mulki2019hsab} are two publicly available corpora for abusive hate speech detection. The first one is in the Tunisian dialect, combining 6K comments. The second one is in Levantine dialect (Syrian, Lebanese, Palestinian, and Jordanian dialects) containing around 6K tweets. These documents are labeled as Abusive, Hate, or Normal.

\cite{mubarak2020arabic} looked at MSA and four major dialects (Egyptian, Levantine, Maghrebi, and Gulf). It presented a systematic method for building an Arabic offensive language tweet dataset that does not favor specific dialects, topics, or genres with 10K tweets. For tweet labeling, they used the count of positive and negative terms based on a polarity lexicon. FastText and Skip-Gram (AraVec skip-gram, Mazajak skip-gram); and deep contextual embeddings, namely BERTbase-multilingual and AraBERT are used. They evaluated different models: SVM, AdaBoost, and Logistic regression.

\cite{mulki2021let} introduced the first Arabic Levantine Twitter dataset for Misogynistic language (LeT-Mi) to be a benchmark dataset for automatic detection of online misogyny written in the Arabic and Levantine dialect. The proposed dataset consists of 6.5K tweets annotated either as neutral (misogynistic-free) or as one of seven misogyny categories: discredit, dominance, cursing/damning, sexual harassment, stereotyping and objectification, derailing, and threat of violence. They used BOW + TF-IDF, SOTA, LSTM, BERT, and Majority class as classifiers.

\cite{duwairi2021deep} investigated the ability of CNN, CNN-LSTM, and BiLSTM-CNN deep learning networks to classify or discover hateful content posted on social media. These deep networks were trained and tested using the ArHS dataset, which consists of around 10K tweets that were annotated to suit hateful speech detection in Arabic. Three types of experiments are reported: first, the binary classification of tweets into Hate or Normal. Ternary classification of tweets into (Hate, Abusive, or Normal), and multi-class classification of tweets into (Misogyny, Racism, Religious Discrimination, Abusive, and Normal).

\cite{aldjanabi2021arabic} have built an offensive and hate speech detection system using a multi-task learning (MTL) model built on top of a pre-trained Arabic language model. The Arabic MTL model was experimented with two different language models to cover MSA and dialect Arabic. They evaluated a new pre-trained model 'MarBERT' to classify both dialect and MSA tweets. They propose a model to explore multi-corpus-based learning using Arabic LMs and MTL to improve the classification performance.

\cite{haidar2017multilingual} presented a solution for the issue of cyberbullying in both Arabic and English languages. The proposed solution is based on machine learning algorithms using a dataset from Lebanon, Syria, the Gulf Area, and Egypt. That dataset contained 35K Arabic texts. In this research, Naïve Bayes and SVM models were chosen to classify the text. The SVM model achieved greater precision.

\cite{abdelali2016farasa} The authors built a large dataset that consists of offensive Arabic words from different dialects and topics. The tweets were labeled into one of these categories: offensive, vulgar, hate speech, or clean. Since the offensive tweets involve implicit insults, the hate speech category was the tweets that contain racism, religious, and ethnic words. Different classifiers were employed in this study; the SVM model with a radial function kernel was mainly used with lexical features and pre-trained static embedding, while Adaptive Boosting and Logistic regression classifiers were employed when using Mazajak embedding. SVM gave the best precision.
 
\textit{According to this literature analysis, we detect that the topic of hate speech detection in the Algerian dialect is not widely considered, and only few works deal with this problem. Furthermore, a lack of Algerian datasets prepared for hate speech is found. All these findings motivate our proposal.}

\section{Our Methodology}
To identify hate speech in messages written in Algerian dialects—whether in Arabic or Latin script— we outline a comprehensive methodology encompassing (1) data gathering, (2) data annotation, (3) feature extraction, (4) model development, and (5) model evaluation and inference. We'll delve into each of these stages in the subsequent sections.

\subsection{Data Collection}
Data collection serves as the foundational step in our approach. To effectively train our models, we require a robust dataset in the Algerian Arabic dialect. To achieve this, we sourced our data from three distinct social networks spanning the years 2017 to 2023:

\textbf{1. YouTube}: Numerous Algerian channels have emerged on YouTube, dedicated to discuss various topics, including politics, religion, social issues, youth concerns, education, and more. We have identified and focused on the most influential ones with a significant following and engagement. We employ the YouTube Data API through a Python script to gather comments from various videos.

\textbf{2. Twitter}: Even if Algerian citizens do not widely use Twitter, we targeted it to collect tweets. We used a list of keywords to search for tweets. Many hashtags were launched between 2017 and 2023 about some situations and crises in Algeria, which enhanced the activity on Twitter, like \RL{ما تشريش زيت ربراب} (do not buy oil of rebrab), \RL{يتنحاو قاع} (remove them all), \RL{العصابة} (mafia),  \RL{لا للعهدة الخامسة } (no for fifth presidential term), etc. During this activity, we used these hashtags to collect an important number of tweets. Two techniques have been used for this objective: (1) Using Twitter API: Until February 2023, we were able to use this API for free and gather tweets. (2) Since February 2023, this API has become paid. Consequently, we used other solutions based on scrapping using the SNScrape library.

\textbf{3. Facebook}: To gather data from Facebook, we selected public pages talking and sharing content about politics, Algerian products, pages of some influencers, mobile operators, etc. We collected the posts, comments, and replies from these various pages. To collect data, we used different solutions: (1) Between 2017 and 2018, we were able to collect data from any public page using Graph API. (2) Since 2019, we have used either FacePager free application to collect data from public pages or (3) Facebook-scraper library for scraping.

From these sources, we have collected more than 2 million documents (messages) in different languages: Arabic, French, English, dialect, etc. The next step consists of filtering only documents written in Algerian dialects, either in Arabic or Latin characters. This work was done manually by a group of collaborators. At the end, we obtained around 900K documents.

\subsection{Data Annotation (Data Labeling)}
To annotate data, we followed two approaches: automatic and manual. We have decided to annotate only the dialect written in Arabic characters. Our approach consists of building one model that detects hate speech only for Algerian dialects written in Arabic characters. Then, a transliteration function is developed to transliterate any Algerian document written in Latin characters into Arabic ones, then use the built model to classify it. For example, "ma tech-rich zit el aliha" becomes "\RL{ما تشريش زيت الالهة}" which means "Don't buy the oil of the gods," which expresses the expensiveness of this oil.
\\
We used a binary annotation: 0 expressing NON-HATE, which represents a document that doesn't contain any hateful or offensive word. 1 in case of a Hateful message containing any hateful word or the meaning and the semantics of the message expresses it.
\\
\textbf{1- Automatic annotation}: For automatic annotation, we prepared a set of hateful keywords in the Algerian dialect discovered from our corpus. These words express the hate and the violence in Algerian speech. This list contains 1.298 words. This list of keywords has been used in a Python script to automatically tag a document with 1 if it contains at least one hateful keyword. In the other case, it is considered as 0. The automatically annotated corpus contains 200K Algerian documents written in Arabic characters.
\\
\textbf{2- Manual annotation}: The automatically annotated documents have been validated manually. A group of annotators checked the annotated corpus and corrected the wrong-labeled documents. The manual step validates 5.644 documents considered for the next step.
\\
\textbf{3- Dataset Augmentation for Enhanced Balance}: To bolster our dataset and enhance its equilibrium, we employed a strategy involving the incorporation of positively labeled subsets sourced from sentiment analysis datasets. In doing so, we reclassified these subsets as non-hateful, under the reasonable assumption that expressions of positive sentiment inherently exclude hate speech. Specifically, we leveraged the dataset available at \url{https://www.kaggle.com/datasets/djoughimehdi/algerian-dialect-review-for-sentiment-analysis}, selecting solely the instances characterized by positive sentiment and relabeling them as 'normal.' However, due to preprocessing constraints, this process yielded a reduced set of just 500 documents.
\\
Moreover, we used the corpus shared by \cite{boucherit2022offensive} containing 8.7K documents in the Algerian dialect. This dataset is labeled manually as Offensive (3.227), Abusive (1.334), and Normal (4.188). We changed the labels of this corpus into Hateful (1) for fused Offensive and Abusive ones and Non-Hateful (0) for Normal ones. This corpus has been filtered and treated to keep 7.345 labeled documents.
\\
At the end of this step, we obtained an annotated balanced corpus of 13.5K documents in Algerian dialect written in Arabic characters, which will be used later to build classifiers.

\subsection{Data Preprocessing}
\begin{figure}
  \centering
  \includegraphics[width=.47\textwidth]{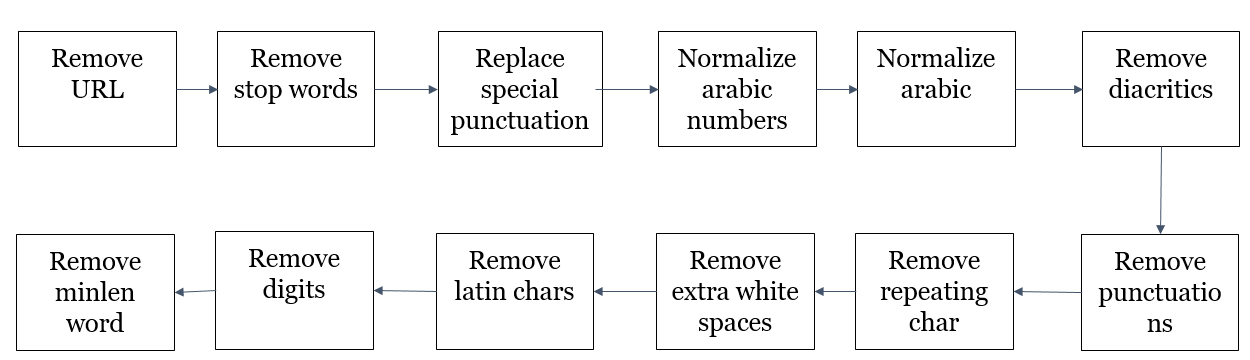}
  \caption{Preprocessing pipeline}
  \label{Fig:preproc}
\end{figure}

Before using any dataset, a cleaning or preprocessing task should be performed. We have defined a set of functions orchestrated in a pipeline, as illustrated in Figure \ref{Fig:preproc}.

\begin{itemize}
\item Remove URL: All URLs in a document are deleted.
\item Remove stop words: The list of Arabic stop words provided by NLTK is used to clean meaningless words. This list has been enriched by a set of stop words detected in the Algerian dialect.
\item Replace special punctuation: some punctuation concatenation can represent meaning and have an added value for the model, like: :) Means happy, :( Means upset, etc. This kind of punctuation is transformed into the corresponding emoji.
\item Normalize Arabic numbers: Arabic numbers are transformed into the classic digits in order to standardize the writing, like \RL{١} = 1, \RL{٢}=2, etc.
\item Normalize Arabic: some letters have special symbols, which needs some treatment. Like:\RL{ ى"="ي" "گ ="ك}
\item Remove diacritics: like the vowel marks \RL{' ّ'} representing Tashdid; \RL{' َ'} meaning Fatha, etc.
\item Remove punctuation: All punctuation except the ones representing emotions are deleted.
\item Remove repeating character: Any repeated character is removed, keeping just one occurrence.
\item Remove extra white spaces: All extra white spaces are deleted.
\item Remove Latin chars: Latin characters in Arabic text are removed to avoid incoherence.
\item Remove digits: In the Arabic dialect, digits do not have any added value. They are removed
\item Remove min-length words: Some words written in less than two positions are deleted. The experiences show that these words are meaningless.
\end{itemize}

The preprocessing of the data was meticulously crafted to cater to the unique characteristics of the Algerian dialect text. By applying rigorous preprocessing to the dialect, the data was made consistent and well-suited for training our models. These preprocessing steps were vital in ensuring that the model was sensitive to the nuances of the language and could effectively classify hate and non-hate content.

\subsection{Data Splitting}
In the development of the models for binary classification of hate speech in the Algerian dialect, our corpus was loaded from CSV file and was stratified before it was split into three distinct sets: training (80\%), validation (10\%), and testing (10\%). The stratification is based on the label column to maintain a balanced representation of hate (1) and non-hate (0) content in each subset. The training set facilitated the model training, while the validation set allowed for unbiased model evaluation during training to prevent over-fitting. The testing set served as an objective assessment of the model's generalization performance beyond the training data. Tokenization, which includes padding and truncation, is also performed. 

\subsection{Model Development}
In this work, we evaluated many classifiers from machine
and deep learning. Below, we discuss the architecture,
methodology, and performance of each model. 
\\
\\
\textbf{1. Linear Support Vector Classifier (LinearSVC):}
We utilized a Linear Support Vector Classifier (LinearSVC) model to investigate how a traditional machine learning approach would perform on this task. The TF-IDF (Term Frequency-Inverse Document Frequency) method was employed to convert the text data into a numerical format suitable for machine learning models. The model was initialized with default parameters and trained on the feature matrix obtained from the TF-IDF vectorization. 
\\
\\
\textbf{2. Gzip + KNN:}
Deep Neural Networks are potent learners capable of tackling a wide array of tasks. However, for relatively straightforward tasks like topic classification they often prove excessive due to their substantial data requirements, high computational demands, and the need for meticulous hyper-parameter tuning. This part of the research centers on a more straightforward alternative known as "compressor-brd text classification," requiring no training parameters. which, despite its astonishing simplicity, exhibits interesting results. 
The approach comprises three key components: (1) utilization of a conventional lossless compression algorithm (gzip in this study); (2) application of the compressor-brd distance metric (Normalized Compression Distance in this study); (3) Implementation of a traditional KNN classifier.
\\
\\

\textbf{3. LSTM \& BiLSTM with Dziri FastText:}
LSTM and BiLSTM are one of the deep learning models that are suitable for NLP problems, mainly in text classification like sentiment analysis and even for hate speech detection. In this paper, we have tested these two models against our corpus. To learn the semantics and context of messages, we used FastText as a word embedding model. In our case, we fine tuned a Dziri FastText model. This later was trained on a huge dataset of Algerian messages in Arabic characters based on the Skip-gram model. 
The obtained model (Dziri FastText) is used to generate an embedding matrix for our built corpus of hate speech. The sequential architecture is composed of:
(i) Embedding layer which is the input layer representing the embedding matrix; (ii) Dropout layer with a rate of 0.2 to prevent over-fitting; (iii) LSTM or Bidirectional LSTM layer with units=100, dropout=0.4, recurrent\_dropout=0.2; (iv) Dropout layer with a rate of 0.2 to prevent over-fitting; (v) Output dense layer, using sigmoid as an activation function.
As optimizer we used Adam, and we used binary crossentropy as a loss function, batch\_size = 64 and epochs= 100.
\\
\\
\textbf{4. Dziribert-FT-HEAD:}
Pre-trained transformers, like BERT, have become the standard in Natural Language Processing due to their exceptional performance in various tasks and languages. The authors in \cite{abdaoui2021dziribert} collected over one million Algerian tweets and developed DziriBERT, the first Algerian language model, outperforming existing models, especially for the Latin script (Arabizi). This demonstrates that a specialized model trained on a relatively small dataset can outshine models trained on much larger datasets. The authors have made DziriBERT\footnote{\url{https://huggingface.co/alger-ia/dziribert}} publicly available to the community.
\\
In this experiments we fine-tuned Dziribert, by incorporating a classification head while keeping the rest of the Dziribert parameters frozen. The classification head consists of three key components: a fully connected layer with 128 units, followed by batch normalization for stability, a dropout layer to mitigate overfitting, and a final fully connected layer that produces a single output value. We apply a sigmoid activation function to ensure the output falls between 0 and 1, which suits our binary classification task. Training employed the binary cross-entropy loss function and the Adam optimizer with a fixed learning rate of 1e-3.
Additionally, a learning rate scheduler was employed to dynamically adjust the learning rate during training for improved convergence.
\\
\\
\textbf{5. DZiriBert with Peft+LoRA:}
In our experiment, we fine-tuned the pre-trained model “DZiriBERT” using techniques called Peft (Parameter-Efficient Fine-Tuning) \cite{peft} + LoRA \cite{hu2021lora}. These methodologies allowed us to tailor the model specifically for the Algerian dialect, making it sensitive to the unique nuances of this language. The Peft configuration is established using the LoRa technique. Parameters such as the reduction factor, scaling factor, dropout rate, and bias are defined according to the task requirements. 
\\
\textit{Peft and LoRa Configuration:} PEFT method has recently emerged as a powerful approach for adapting large-scale pre-trained language models (PLMs) to various downstream applications without fine-tuning all the model's parameters. Given that fine-tuning such models can be prohibitively costly, PEFT offers a viable alternative by only fine-tuning a small number of (extra) model parameters. This greatly decreases the computational and storage costs without compromising performance. 
\\
LoRA is a technique specifically designed to make the fine-tuning of large models more efficient and memory-friendly. The essential idea behind LoRA is to represent weight updates using two smaller matrices (referred to as update matrices) through a low-rank decomposition. While the original weight matrix remains frozen, these new matrices are trained to adapt to the new data, keeping the overall number of changes minimal. LoRA has many advantages, mainly the: (1) Efficiency: by significantly reducing the number of trainable parameters, LoRA makes fine-tuning more manageable. (2) Portability: Since the original pre-trained weights are kept frozen, multiple lightweight LoRA models can be created for various downstream tasks. (3) Performance: LoRA achieves performance comparable to fully fine-tuned models without adding any inference latency. (4) Versatility: Though typically applied to attention blocks in Transformer models, LoRA's principles can, in theory, be applied to any subset of weight matrices in a neural network.
\\
\textit{Model Initialization:} DZiriBERT is loaded and configured with Peft using the defined parameters. The model is then fine-tuned using the tokenized datasets. We configure our model using the LoraConfig class, which includes the following hyperparameters:
\\- Task Type: We set the task type to Sequence Classification (SEQ\_CLS), where the model is trained to map an entire sequence of tokens to a single label.
Target Modules: The target modules are set to "query" and "value".
\\- Rank (r): We employ a low-rank approximation with a rank =16 for the LoRA matrices.
\\- Scaling Factor ($\alpha$): The LoRA layer utilizes a scaling factor=32, which serves as a regularization term.
\\- Dropout Rate: We introduce a dropout rate of 0.35 in the LoRA matrices to improve generalization.
\\- Bias: The bias term is set to "none," reducing the model complexity.
\\
\textit{Training Process}: The model is trained using custom training arguments, including learning rate, batch sizes, epochs, and evaluation strategies. The training process leverages the Hugging Face Trainer class, providing a streamlined approach to model fine-tuning. We train our model with the following parameters:
\\-learning\_rate=1e-3: Specifies the learning rate as 1e-3. Learning rate controls how quickly or slowly a model learns during the training process.
\\-per\_device\_train\_batch\_size=16: This indicates that each device used for training (usually a GPU) will handle a batch of 16 samples during each training iteration.
\\- per\_device\_eval\_batch\_size=32: Similar to the above, but for evaluation, each device will process batches of 32 samples.
\\- num\_train\_epochs=5: The training process will go through the entire training dataset 5 times. An epoch is one complete forward and backward pass of all the training examples.
\\- weight\_decay=0.01: This is a regularization technique that helps prevent the model from fitting the training data too closely (overfitting). A weight decay of 0.01 will be applied.
\\- evaluation\_strategy="epoch": Evaluation will be performed at the end of each epoch. This allows you to check the performance of your model more frequently and make adjustments if needed.
\\- save\_strategy="epoch": The model will be saved at the end of each epoch, allowing you to revert to the model's state at the end of any given epoch if necessary.
\\- load\_best\_model\_at\_end=True: Once all training and evaluation are completed, the best-performing model will be loaded back into memory. This ensures that you always have access to the best model when your training is complete.
\\
\\
\textbf{6. Dzarashield:} 
We built the Dzarabert\footnote{\url{https://huggingface.co/Sifal/dzarabert}} which is a modification of the original Dziribert model that involves pruning the embedding layer, specifically removing tokens that contain non-Arabic characters. This pruning significantly reduces the number of trainable parameters, resulting in faster training times and improved inference speed for the model. This approach is aimed at optimizing the model's performance for tasks involving Arabic-based text while minimizing unnecessary complexity and computational overhead.
Dzarashield\footnote{\url{https://huggingface.co/Sifal/dzarashield}} is built upon the Dzarabert base model by incorporating a classification head. This classification head consists of sequential architecture including: a linear layer (input: 768, output: 768), followed by a Rectified Linear Unit (ReLU) activation function; a dropout layer (dropout rate: 0.1); and another linear layer (input: 768, output: 2) for binary classification. The model's hyperparameters were determined through experimentation: a learning rate (lr) of 1.3e-05, a batch size of 16, and training for 4 epochs. The Adam optimizer was used with its default parameters for optimization during training. 
Experimentation resulted in a better score when updating all the weights of the model rather than freezing the base BERT model and updating the classification head.
\\
\\
\textbf{7. Multilingual E5 Model:}
We conducted a fine-tuning process on a pre-existing model, specifically the Multilingual E5 base model \cite{wang2022text}. Our primary objective was to ascertain the efficacy of a multilingual model within the context of the Algerian dialect. In adherence to the training methodology, the prefix "query:'' was systematically introduced to each data row. This precautionary measure was deemed necessary \cite{wang2022text} to avert potential indications of performance deterioration that might arise in the absence of such preprocessing. 
The foundation of our investigation rested upon the initialization of the pre-trained base model using the xlm-roberta-base\footnote{\url{https://huggingface.co/xlm-roberta-base}} architecture, which was trained on a mixture of multilingual datasets. The model is fine-tuned with an additional Dense layer followed by a Dropout Layer. The model is trained with custom hyperparameters for fine-tuning (Warmup Steps: 100; Weight Decay: 0.01 ; Epoch: 5 ; Probability of Dropout: 0.1; Train batch size: 16 ; Evaluation batch size: 64)
\\
\\
\textbf{8. sbert-distill-multilingual Fine Tuned:}
Similar to the Multilingual E5 Model, we fine-tuned a pre-trained model known as sbert-distil-multilingual model from sentence transformer to investigate how well a multilingual model performs in Algerian Dialect. The pre-trained model is based on a fixed (monolingual) teacher model that produces sentence embeddings with our desired properties in one language. The student model is supposed to mimic the teacher model, i.e., the same English sentence should be mapped to the same vector by the teacher and by the student model. The model is fine-tuned with an additional Dropout layer and a GeLU layer via K-Fold cross validation.
The model is trained with custom hyperparameters for fine-tuning (Warmup Steps: 100; Weight Decay: 0.01; Probability of Dropout: 0.1 ; Epoch: 10 ; K-Fold: 4 ; Train batch size: 16 ; Evaluation batch size: 64)
\\
\\
\textbf{9 AraT5v2-HateDetect}
AraT5-base is the result of testing the T5 model (mT5)\footnote{\url{https://huggingface.co/docs/transformers/model_doc/mt5}} on Arabic. For comparison, three robust Arabic T5-style models are pre-trained and evaluated on ARGEN dataset \cite{nagoudi2021arat5}. Surprisingly, despite being trained on approximately 49\% less data, these models outperformed mT5 in the majority of ARGEN tasks, achieving several new state-of-the-art results. 
The AraT5v2-base-1024 model \footnote{\url{https://huggingface.co/UBC-NLP/AraT5v2-base-1024}} introduces several improvements compared to its predecessor, AraT5-base :
\\- More Data: AraT5v2-base-1024 is trained on a larger and more diverse Arabic dataset. This means it has been exposed to a wider range of Arabic text, enhancing its language understanding capabilities.
\\- Larger Sequence Length: This version increases the maximum sequence length from 512 to 1024. This extended sequence length allows the model to handle longer texts, making it more versatile in various NLP tasks.
\\- Faster Convergence: During the fine-tuning process, AraT5v2-base-1024 converges approximately 10 times faster than the previous version (AraT5-base). This can significantly speed up the training and fine-tuning processes, making it more efficient.
\\- Extra IDs: AraT5v2-base-1024 supports 100 sentinel tokens, also known as unique mask tokens. This allows for more flexibility and customization when using the model for specific tasks.
\\
Overall, these enhancements make AraT5v2-base-1024 a more powerful and efficient choice for Arabic natural language processing tasks compared to its predecessor, and it is recommended for use in place of AraT5-base.
AraT5v2-HateDetect\footnote{\url{https://huggingface.co/Sifal/AraT5v2-HateDetect}} is a fine-tuned model based on AraT5v2-base-1024, specifically tailored for the hate detection task. The fine-tuning process involves conditioning the decoder's labels, which include target input IDs and target attention masks, based on the encoder's source documents, which consist of source input IDs and source attention masks. After experimentation, the following hyperparameters were chosen for training AraT5v2-HateDetect (Training Batch Size: 16; Learning Rate: 3e-5; Number of Training Epochs: 4).
These hyperparameters were determined to optimize the model's performance on the hate detection task. The chosen batch size, learning rate, and training epochs collectively contribute to the model's ability to learn and generalize effectively for this specific NLP task.

\subsection{Evaluation and Inference}
To evaluate the different models, we used four main metrics: Accuracy, Precision, F1-Score, and Recall. To classify a message in case where it is written in Arabizi (a specific dialect using Latin characters), a transliteration process was implemented to convert the text into Arabic characters based on lang-trans\footnote{\url{https://pypi.org/project/lang-trans/}} library. 

\section{Experiments and Results}
To train and evaluate our models, we used TensorFlow and Pytorch deep learning frameworks. We used Google Colab and Kaggle GPUs to accelerate the experiments. In table \ref{tab:evolution_of_test_bed}, we will provide the detailed results that we obtained.

\begin{table}
	\caption{The results of each model (FT: Fine Tuned}
	\centering
	\begin{tabular}{lllll}
		\toprule
		
		Model Name & Accuracy & Precision & Recall & F1 Score  \\
		\midrule
		LinearSVC & 0.83 & 0.84(Class0); 0.72(Class1)  & 0.96(Class0); 0.36 (Class1) & 0.9(Class0); 0.48 (Class1) \\    
   gzip + KNN & 0.67 & 0.63  & 0.56  & 0.60     \\    
   Dziribert-FT-HEAD     & 0.83 & 0.81  & 0.81  & 0.81      \\
   LSTM     & 0.70 & 0.61  & 0.75  & 0.67     \\
   Bidirect LSTM & 0.68 & 0.59  & 0.81  & 0.68     \\
   DZiriBERT FT PEFT+LoRA & 0.86 & 0.83  & 0.85  & 0.84 \\
   Multilingual-E5-base FT    & 0.84 & 0.8  & 0.81  & 0.80     \\
   sbert-distill-multilingual FT & 0.80 & 0.74  & 0.81  & 0.77     \\
   DzaraShield & 0.87 & 0.87  & 0.87  & 0.87     \\
   AraT5v2-HateDetect & 0.84 & 0.83  & 0.84  & 0.83    \\
		\bottomrule
	\end{tabular}
	\label{tab:evolution_of_test_bed}
\end{table}

\textbf{\textit{Linear Support Vector Classifier (LinearSVC)}}: The LinearSVC model offered a competitive accuracy but struggled with the recall for the hate speech class. The precision and recall trade-off indicates possible challenges in differentiating between the subtle nuances of hate and non-hate speech in the dialect. The model exhibited high precision and recall for class 0 but showed room for improvement for class 1, particularly in terms of recall. This suggests that while the model is quite good at identifying class 0, it could be improved for identifying class 1. 

\textbf{\textit{gzip + KNN}}: One of the worst models in terms of capabilities, although it is diverging from the baseline it is unclear whether these results will hold in out of distribution cases, especially when we know that there is no underlying process in the model that captures  semantic representations of the documents.

\textbf{\textit{Dziribert-FT-HEAD:}} the model exhibits a noteworthy precision score, signifying its accuracy in correctly classifying instances as hate speech or not. However, the relatively lower recall score suggests that it missed identifying some hate speech instances. This discrepancy might be attributed to the model's lack of specialized handling for the nuances of the Algerian dialect, potentially causing it to overlook certain hate speech patterns unique to that context.\\
Despite this, the model's overall accuracy remains commendably high, indicating its robust performance in making accurate predictions. Additionally, the balanced precision and recall values underline its ability to strike a reasonable trade-off between minimizing false positives and false negatives, a crucial aspect in hate speech detection.\\
The F1 Score, being the harmonic mean of precision and recall, further validates the model's capacity to effectively identify positive samples while avoiding misclassification of negative ones. The model consistently demonstrates strong performance across multiple evaluation metrics, especially in terms of accuracy and F1 score. These results reaffirm the practicality and effectiveness of employing deep learning techniques for the challenging task of hate speech detection.

\textbf{\textit{LSTM and BiLSTM with FastText-DZ:}} Unfortunately, the results of this model are among the worst ones. The literature shows the strength of LSTM and BiLSTM in this kind of NLP project, but this is not the case for this project. The low precision is due to the incapability of the model to classify correctly the hate class. FastText is a good word embedding model that captures the context and semantics of a document. However, in this case, it does not perform well because of the fine-tuning done where we took an Arabic FastText and fine-tune it on Algerian dataset written in Arabic characters.

\textbf{\textit{DZiriBert with Peft+LoRA:}} We utilize both PEFT and LoRA to fine-tune DZiriBERT, a model specifically adapted to the Algerian dialect. By employing these techniques, we were able to create a highly effective and efficient model for hate speech detection in the Algerian dialect while keeping computational costs at a minimum.

\textbf{\textit{Multilingual-E5-base Fine Tuned and sbert-distill-multilingual Fine Tuned}}: The outcomes obtained from these models are noteworthy; nonetheless, their performances pale when compared with the parameter-efficient fine-tuning on the DZiriBERT model.

\textbf{\textit{DzaraShield}}: The results returned by this model are satisfying considering the relatively low quantity of data it was finetuned on, this exhibits further that the pretraining plays the major role on downstream takes such as classification in our case, especially that the base model is an encoder only architecture which  captures contextual information from the input data, making it useful for a wide range of text classification tasks. 

\textbf{\textit{AraT5v2-HateDetect}}: The results are slightly inferior to Dzarashield. One possible explanation is the increased complexity of the architecture when compared to the Dzarabert base model. Consequently, fine-tuning becomes a more intricate task due to the larger hyperparameter search space and the limited resources in terms of computing power and data availability. As a result, it is reasonable to expect that these models would perform similarly in real-world scenarios.

\subsection{Results Discussion}
The DzaraShield model has demonstrated remarkable capability in detecting hate speech in the Algerian dialect. Its outstanding precision score highlights its reliability in accurately identifying instances of hate speech. Additionally, it maintains a balanced precision and recall, indicating that it does not excessively sacrifice precision to achieve its higher recall. Such a balanced model holds considerable advantages, particularly when both false positives and false negatives carry significant consequences.

For the other models, mainly LSTM or BiLSTM with Dziri FastText, more fine-tuning should be performed to enhance the results. Moreover, future work may include hyperparameter tuning, class balancing techniques, or the integration of more complex models to improve performance across both classes.

The disparity between precision and recall in certain models warrants further investigation. Delving deeper into this issue could yield valuable insights into specific aspects of the dialect that might be contributing to this imbalance. Future experiments should prioritize understanding and addressing these discrepancies, with the goal of enhancing recall without compromising precision.

The results from various experimental models underscore the intricacies involved in hate speech detection in the Algerian dialect. While traditional machine learning and deep learning approaches provided some valuable insights, they fell short in capturing the dialect's nuanced characteristics. In contrast, the DzaraShield model emerged as the most successful approach, emphasizing the pivotal role of Encoder-only models in the realm of projects of this nature.

These findings offer valuable insights for future work in this area and underscore the potential of leveraging domain-specific knowledge, advanced fine-tuning techniques, and sophisticated architectures for the effective detection of hate speech in under-studied and complex dialects such as Algerian.

\section{Conclusion}
The importance of hate speech detection on social networks has encouraged many researchers to build solutions (corpora and classifiers) to detect suspect messages. The literature review shows that most works are interested in text in structured languages like English, French, Arabic, etc. However, few works deal with dialects, mainly the Algerian one, which is known for its complexity and variety. To fill in the gap, we propose in this paper a complete NLP approach to detect hate speech in the Algerian dialect. We built an annotated corpus of more than 13,5K documents, which is used to evaluate various deep learning architectures. The obtained results are very promising, where the most accurate was the DzaraShield . 
\\
Looking ahead, there is significant potential to enhance inference speed, particularly for the Dziribert-based and multilingual models. While this project primarily focused on Arabic characters, our next step will be to address the dialect when written in Latin characters. Embracing both Arabic and Latin characters will more accurately capture the nuances of the written Algerian dialect. Finally, we plan to expand our corpus size and explore alternative deep-learning architectures.

\section{Acknowledgments}
We would like to thank every person who has contributed to this project: Micha Freidin, Viktor Ivanenko, Piyush Aaryan, Yassine Elboustani, Tasneem Elyamany, Cephars Bonacci, Nolan Wang and Lydia Khelifa Chibout. We would also like to thank Omdena organization for giving us this valuable opportunity.

\bibliographystyle{unsrtnat}
\bibliography{references}

\begin{thebibliography}{43}
\providecommand{\natexlab}[1]{#1}
\providecommand{\url}[1]{\texttt{#1}}
\expandafter\ifx\csname urlstyle\endcsname\relax
  \providecommand{\doi}[1]{doi: #1}\else
  \providecommand{\doi}{doi: \begingroup \urlstyle{rm}\Url}\fi

\bibitem[MacAvaney et~al.(2019)MacAvaney, Yao, Yang, Russell, Goharian, and Frieder]{macavaney2019hate}
Sean MacAvaney, Hao-Ren Yao, Eugene Yang, Katina Russell, Nazli Goharian, and Ophir Frieder.
\newblock Hate speech detection: Challenges and solutions.
\newblock \emph{PloS one}, 14\penalty0 (8):\penalty0 e0221152, 2019.

\bibitem[Bogdani et~al.(2021)Bogdani, Faloppa, and Karaj]{bogdani2021beyond}
Mirela~P Bogdani, Federico Faloppa, and Xheni Karaj.
\newblock Beyond definitions. a call for action against hate speech in albania. a comprehensive study november 2021.
\newblock 2021.

\bibitem[Nockleby(2000)]{nockleby2000hate}
JT~Nockleby.
\newblock hate speech in encyclopedia of the american constitution. electronic journal of academic and special librarianship.
\newblock 2000.

\bibitem[Guellil et~al.(2022)Guellil, Adeel, Azouaou, Boubred, Houichi, and Moumna]{guellil2022ara}
Imane Guellil, Ahsan Adeel, Faical Azouaou, Mohamed Boubred, Yousra Houichi, and Akram~Abdelhaq Moumna.
\newblock Ara-women-hate: An annotated corpus dedicated to hate speech detection against women in the arabic community.
\newblock In \emph{Proceedings of the Workshop on Dataset Creation for Lower-Resourced Languages within the 13th Language Resources and Evaluation Conference}, pages 68--75, 2022.

\bibitem[Davidson et~al.(2017)Davidson, Warmsley, Macy, and Weber]{davidson2017automated}
Thomas Davidson, Dana Warmsley, Michael Macy, and Ingmar Weber.
\newblock Automated hate speech detection and the problem of offensive language.
\newblock In \emph{Proceedings of the international AAAI conference on web and social media}, volume~11, pages 512--515, 2017.

\bibitem[Fortuna and Nunes(2018)]{fortuna2018survey}
Paula Fortuna and S{\'e}rgio Nunes.
\newblock A survey on automatic detection of hate speech in text.
\newblock \emph{ACM Computing Surveys (CSUR)}, 51\penalty0 (4):\penalty0 1--30, 2018.

\bibitem[Schmidt and Wiegand(2017)]{schmidt2017survey}
Anna Schmidt and Michael Wiegand.
\newblock A survey on hate speech detection using natural language processing.
\newblock In \emph{Proceedings of the fifth international workshop on natural language processing for social media}, pages 1--10, 2017.

\bibitem[Battistelli et~al.(2020)Battistelli, Bruneau, and Dragos]{battistelli2020building}
Delphine Battistelli, Cyril Bruneau, and Valentina Dragos.
\newblock Building a formal model for hate detection in french corpora.
\newblock \emph{Procedia Computer Science}, 176:\penalty0 2358--2365, 2020.

\bibitem[Alkomah and Ma(2022)]{alkomah2022literature}
Fatimah Alkomah and Xiaogang Ma.
\newblock A literature review of textual hate speech detection methods and datasets.
\newblock \emph{Information}, 13\penalty0 (6):\penalty0 273, 2022.

\bibitem[Plaza-del Arco et~al.(2021)Plaza-del Arco, Molina-Gonz{\'a}lez, Urena-L{\'o}pez, and Mart{\'\i}n-Valdivia]{plaza2021comparing}
Flor~Miriam Plaza-del Arco, M~Dolores Molina-Gonz{\'a}lez, L~Alfonso Urena-L{\'o}pez, and M~Teresa Mart{\'\i}n-Valdivia.
\newblock Comparing pre-trained language models for spanish hate speech detection.
\newblock \emph{Expert Systems with Applications}, 166:\penalty0 114120, 2021.

\bibitem[Albadi et~al.(2018)Albadi, Kurdi, and Mishra]{albadi2018they}
Nuha Albadi, Maram Kurdi, and Shivakant Mishra.
\newblock Are they our brothers? analysis and detection of religious hate speech in the arabic twittersphere.
\newblock In \emph{2018 IEEE/ACM International Conference on Advances in Social Networks Analysis and Mining (ASONAM)}, pages 69--76. IEEE, 2018.

\bibitem[Mulki et~al.(2019)Mulki, Haddad, Ali, and Alshabani]{mulki2019hsab}
Hala Mulki, Hatem Haddad, Chedi~Bechikh Ali, and Halima Alshabani.
\newblock L-hsab: A levantine twitter dataset for hate speech and abusive language.
\newblock In \emph{Proceedings of the third workshop on abusive language online}, pages 111--118, 2019.

\bibitem[Mezzoudj et~al.(2019)Mezzoudj, Loukam, and Belkredim]{mezzoudj2019arabic}
Fr{\'e}ha Mezzoudj, Mourad Loukam, and Fatma~Zohra Belkredim.
\newblock Arabic algerian oranee dialectal language modelling oriented topic.
\newblock \emph{International Journal of Informatics and Applied Mathematics}, 2\penalty0 (2):\penalty0 1--14, 2019.

\bibitem[Boucherit and Abainia(2022)]{boucherit2022offensive}
Oussama Boucherit and Kheireddine Abainia.
\newblock Offensive language detection in under-resourced algerian dialectal arabic language.
\newblock \emph{arXiv preprint arXiv:2203.10024}, 2022.

\bibitem[Menifi et~al.(2022)Menifi, Moussa, and Mazari]{menifi2022transfer}
Djamila Menifi, Wiam Moussa, and Ahmed~Cherif Mazari.
\newblock \emph{Transfer Learning and Deep Learning for Multilingual Algerian Dialect Hate Speech Detection}.
\newblock PhD thesis, 2022.

\bibitem[Abdelli et~al.(2019)Abdelli, Guerrouf, Tibermacine, and Abdelli]{abdelli2019sentiment}
Adel Abdelli, Fay{\c{c}}al Guerrouf, Okba Tibermacine, and Belkacem Abdelli.
\newblock Sentiment analysis of arabic algerian dialect using a supervised method.
\newblock In \emph{2019 International Conference on Intelligent Systems and Advanced Computing Sciences (ISACS)}, pages 1--6. IEEE, 2019.

\bibitem[Guellil et~al.(2021)Guellil, Adeel, Azouaou, Boubred, Houichi, and Moumna]{guellil2021sexism}
Imane Guellil, Ahsan Adeel, Faical Azouaou, Mohamed Boubred, Yousra Houichi, and Akram~Abdelhaq Moumna.
\newblock Sexism detection: The first corpus in algerian dialect with a code-switching in arabic/french and english.
\newblock \emph{arXiv preprint arXiv:2104.01443}, 2021.

\bibitem[De~Gibert et~al.(2018)De~Gibert, Perez, Garc{\'\i}a-Pablos, and Cuadros]{de2018hate}
Ona De~Gibert, Naiara Perez, Aitor Garc{\'\i}a-Pablos, and Montse Cuadros.
\newblock Hate speech dataset from a white supremacy forum.
\newblock \emph{arXiv preprint arXiv:1809.04444}, 2018.

\bibitem[Al-Hassan and Al-Dossari(2019)]{al2019detection}
Areej Al-Hassan and Hmood Al-Dossari.
\newblock Detection of hate speech in social networks: a survey on multilingual corpus.
\newblock In \emph{6th international conference on computer science and information technology}, volume~10, pages 10--5121, 2019.

\bibitem[Boudad et~al.(2018)Boudad, Faizi, Thami, and Chiheb]{boudad2018sentiment}
Naaima Boudad, Rdouan Faizi, Rachid Oulad~Haj Thami, and Raddouane Chiheb.
\newblock Sentiment analysis in arabic: A review of the literature.
\newblock \emph{Ain Shams Engineering Journal}, 9\penalty0 (4):\penalty0 2479--2490, 2018.

\bibitem[Habash(2022)]{habash2022introduction}
Nizar~Y Habash.
\newblock \emph{Introduction to Arabic natural language processing}.
\newblock Springer Nature, 2022.

\bibitem[Sharaf and Atwell(2012)]{sharaf2012qurana}
Abdul-Baquee~M Sharaf and Eric Atwell.
\newblock Qurana: Corpus of the quran annotated with pronominal anaphora.
\newblock In \emph{Lrec}, pages 130--137, 2012.

\bibitem[Abainia et~al.(2022)Abainia, Kara, and Hamouni]{abainia2022new}
Kheireddine Abainia, Kenza Kara, and Tassadit Hamouni.
\newblock A new corpus and lexicon for offensive tamazight language detection.
\newblock In \emph{Proceedings of the 7th International Workshop on Social Media World Sensors}, pages 1--6, 2022.

\bibitem[Mazari and Kheddar(2023)]{mazari2023deep}
Ahmed~Cherif Mazari and Hamza Kheddar.
\newblock Deep learning-based analysis of algerian dialect dataset targeted hate speech, offensive language and cyberbullying.
\newblock \emph{International Journal of Computing and Digital Systems}, 2023.

\bibitem[Guellil et~al.(2020)Guellil, Adeel, Azouaou, Chennoufi, Maafi, and Hamitouche]{guellil2020detecting}
Imane Guellil, Ahsan Adeel, Faical Azouaou, Sara Chennoufi, Hanene Maafi, and Thinhinane Hamitouche.
\newblock Detecting hate speech against politicians in arabic community on social media.
\newblock \emph{International Journal of Web Information Systems}, 16\penalty0 (3):\penalty0 295--313, 2020.

\bibitem[Mohdeb et~al.(2022)Mohdeb, Laifa, Zerargui, and Benzaoui]{mohdeb2022evaluating}
Djamila Mohdeb, Meriem Laifa, Fayssal Zerargui, and Omar Benzaoui.
\newblock Evaluating transfer learning approach for detecting arabic anti-refugee/migrant speech on social media.
\newblock \emph{Aslib Journal of Information Management}, 74\penalty0 (6):\penalty0 1070--1088, 2022.

\bibitem[ALBayari and Abdallah(2022)]{albayari2022instagram}
Reem ALBayari and Sherief Abdallah.
\newblock Instagram-based benchmark dataset for cyberbullying detection in arabic text.
\newblock \emph{Data}, 7\penalty0 (7):\penalty0 83, 2022.

\bibitem[Al-Ajlan and Ykhlef(2018)]{al2018optimized}
Monirah~A Al-Ajlan and Mourad Ykhlef.
\newblock Optimized twitter cyberbullying detection based on deep learning.
\newblock In \emph{2018 21st Saudi Computer Society National Computer Conference (NCC)}, pages 1--5. IEEE, 2018.

\bibitem[Haidar et~al.(2019)Haidar, Chamoun, and Serhrouchni]{haidar2019arabic}
Batoul Haidar, Maroun Chamoun, and Ahmed Serhrouchni.
\newblock Arabic cyberbullying detection: Enhancing performance by using ensemble machine learning.
\newblock In \emph{2019 international conference on internet of things (ithings) and ieee green computing and communications (greencom) and ieee cyber, physical and social computing (cpscom) and ieee smart data (smartdata)}, pages 323--327. IEEE, 2019.

\bibitem[Mubarak et~al.(2017)Mubarak, Darwish, and Magdy]{mubarak2017abusive}
Hamdy Mubarak, Kareem Darwish, and Walid Magdy.
\newblock Abusive language detection on arabic social media.
\newblock In \emph{Proceedings of the first workshop on abusive language online}, pages 52--56, 2017.

\bibitem[Alakrot et~al.(2018)Alakrot, Murray, and Nikolov]{alakrot2018dataset}
Azalden Alakrot, Liam Murray, and Nikola~S Nikolov.
\newblock Dataset construction for the detection of anti-social behaviour in online communication in arabic.
\newblock \emph{Procedia Computer Science}, 142:\penalty0 174--181, 2018.

\bibitem[Haddad et~al.(2019)Haddad, Mulki, and Oueslati]{haddad2019t}
Hatem Haddad, Hala Mulki, and Asma Oueslati.
\newblock T-hsab: A tunisian hate speech and abusive dataset.
\newblock In \emph{International conference on Arabic language processing}, pages 251--263. Springer, 2019.

\bibitem[Mubarak et~al.(2020)Mubarak, Rashed, Darwish, Samih, and Abdelali]{mubarak2020arabic}
Hamdy Mubarak, Ammar Rashed, Kareem Darwish, Younes Samih, and Ahmed Abdelali.
\newblock Arabic offensive language on twitter: Analysis and experiments.
\newblock \emph{arXiv preprint arXiv:2004.02192}, 2020.

\bibitem[Mulki and Ghanem(2021)]{mulki2021let}
Hala Mulki and Bilal Ghanem.
\newblock Let-mi: an arabic levantine twitter dataset for misogynistic language.
\newblock \emph{arXiv preprint arXiv:2103.10195}, 2021.

\bibitem[Duwairi et~al.(2021)Duwairi, Hayajneh, and Quwaider]{duwairi2021deep}
Rehab Duwairi, Amena Hayajneh, and Muhannad Quwaider.
\newblock A deep learning framework for automatic detection of hate speech embedded in arabic tweets.
\newblock \emph{Arabian Journal for Science and Engineering}, 46:\penalty0 4001--4014, 2021.

\bibitem[Aldjanabi et~al.(2021)Aldjanabi, Dahou, Al-qaness, Elaziz, Helmi, and Dama{\v{s}}evi{\v{c}}ius]{aldjanabi2021arabic}
Wassen Aldjanabi, Abdelghani Dahou, Mohammed~AA Al-qaness, Mohamed~Abd Elaziz, Ahmed~Mohamed Helmi, and Robertas Dama{\v{s}}evi{\v{c}}ius.
\newblock Arabic offensive and hate speech detection using a cross-corpora multi-task learning model.
\newblock In \emph{Informatics}, volume~8, page~69. MDPI, 2021.

\bibitem[Haidar et~al.(2017)Haidar, Chamoun, and Serhrouchni]{haidar2017multilingual}
Batoul Haidar, Maroun Chamoun, and Ahmed Serhrouchni.
\newblock A multilingual system for cyberbullying detection: Arabic content detection using machine learning.
\newblock \emph{Advances in Science, Technology and Engineering Systems Journal}, 2\penalty0 (6):\penalty0 275--284, 2017.

\bibitem[Abdelali et~al.(2016)Abdelali, Darwish, Durrani, and Mubarak]{abdelali2016farasa}
Ahmed Abdelali, Kareem Darwish, Nadir Durrani, and Hamdy Mubarak.
\newblock Farasa: A fast and furious segmenter for arabic.
\newblock In \emph{Proceedings of the 2016 conference of the North American chapter of the association for computational linguistics: Demonstrations}, pages 11--16, 2016.

\bibitem[Abdaoui et~al.(2021)Abdaoui, Berrimi, Oussalah, and Moussaoui]{abdaoui2021dziribert}
Amine Abdaoui, Mohamed Berrimi, Mourad Oussalah, and Abdelouahab Moussaoui.
\newblock Dziribert: a pre-trained language model for the algerian dialect.
\newblock \emph{arXiv preprint arXiv:2109.12346}, 2021.

\bibitem[Mangrulkar et~al.(2022)Mangrulkar, Gugger, Debut, Belkada, and Paul]{peft}
Sourab Mangrulkar, Sylvain Gugger, Lysandre Debut, Younes Belkada, and Sayak Paul.
\newblock Peft: State-of-the-art parameter-efficient fine-tuning methods.
\newblock \url{https://github.com/huggingface/peft}, 2022.

\bibitem[Hu et~al.(2021)Hu, Shen, Wallis, Allen-Zhu, Li, Wang, Wang, and Chen]{hu2021lora}
Edward~J Hu, Yelong Shen, Phillip Wallis, Zeyuan Allen-Zhu, Yuanzhi Li, Shean Wang, Lu~Wang, and Weizhu Chen.
\newblock Lora: Low-rank adaptation of large language models.
\newblock \emph{arXiv preprint arXiv:2106.09685}, 2021.

\bibitem[Wang et~al.(2022)Wang, Yang, Huang, Jiao, Yang, Jiang, Majumder, and Wei]{wang2022text}
Liang Wang, Nan Yang, Xiaolong Huang, Binxing Jiao, Linjun Yang, Daxin Jiang, Rangan Majumder, and Furu Wei.
\newblock Text embeddings by weakly-supervised contrastive pre-training.
\newblock \emph{arXiv preprint arXiv:2212.03533}, 2022.

\bibitem[Nagoudi et~al.(2021)Nagoudi, Elmadany, and Abdul-Mageed]{nagoudi2021arat5}
E~Moatez~Billah Nagoudi, A~Elmadany, and M~Abdul-Mageed.
\newblock Arat5: Text-to-text transformers for arabic language understanding and generation.
\newblock \emph{arXiv preprint arXiv:2109.12068}, 2021.

\end{thebibliography}

\end{document}